# A Tractable POMDP for a Class of Sequencing Problems

Paat Rusmevichientong, Benjamin Van Roy
Stanford University
Stanford, CA 94305
{paatrus, bvr}@ stanford.edu

## Abstract

We consider a partially observable Markov decision problem (POMDP) that models a class of sequencing problems. Although POMDPs are typically intractable, our formulation admits tractable solution. Instead of maintaining a value function over a high-dimensional set of belief states, we reduce the state space to one of smaller dimension, in which grid-based dynamic programming techniques are effective. We develop an error bound for the resulting approximation, and discuss an application of the model to a problem in targeted advertising.

## 1 Introduction

Motivated by a problem in targeted advertising, we consider a partially observable Markov decision problem that models a customer's responses to different products that are presented during a marketing campaign. In our model, a customer's responses are dependent random variables, but are assumed to be conditionally independent given another random variable $X$, which can be interpreted as the profile of the customer. The "profile" can consist of both known and unknown characteristics that influence a purchasing decision, e.g. age, gender, or income.

Starting with a prior distribution of $X$, which represents initial beliefs, our model maintains a belief state over a customer's profile. As information is acquired through interaction with the customer, beliefs are updated by computing posterior distributions, which then guide presentation of additional products.

Our formulation constitutes a specialized class of partially observable Markov decision problems (POMDPs). POMDPs are generally intractable [4], and their intractability has motivated the development of approximation methods. It is well known that a POMDP can be viewed as a fully observable Markov decision problem (MDP), for which the state of the MDP corresponds to the posterior distribution over states of the POMDP [1]. Lovejoy [3] has proposed a method that approximates the value function over a grid of points in the space of posterior distributions. A grid is formed; then, an upper and a lower bound for the value function are computed. A policy is generated based on the lower bound. The difference in the performance between this policy and that of the optimal policy is bounded by the gap between the upper and lower approximations. Unfortunately, the number of grid points required for an effective approximation typically grows exponentially in the number of states of the POMDP.

As we will show in this paper, under a further assumption on the form of probabilistic relationships between customer profile and behavior, an approximate solution to the class of POMDPs we consider can be computed efficiently. In particular, for this class of problems, we offer a grid-based dynamic programming method that entails forming a grid on a lower dimensional Euclidean space rather than the space of posterior distributions. The identification of this tractable class of POMDPs together with a solution method constitute the main contribution of this paper.

The paper is organized as follows. In the next section, we provide a general problem formulation. Then, in Section 3, we discuss why standard dynamic programming techniques are computationally too demanding. A class of distributions that admits efficient solution is introduced in Section 4. An effective approximation technique is then developed in Section 5. In Section 6, we consider a possible application of our model to a problem in targeted advertising. Finally, we conclude with a discussion of extensions and future work.

## 2 Problem Formulation

Let $X$ be a random variable taking values in a finite set $\mathcal{S}$ with a prior distribution $\phi_0$ satisfying $\phi_0(x) > 0$ for all $x \in \mathcal{S}$, and let $\mathcal{U}$ be a set of available decisions. The random variable $X$ might correspond to a customer's profile, and $\mathcal{U}$ to a set of products that we can present to the customer during a marketing campaign. Let $Y$ be a binary random variable that denotes the customer's response, with $Y = 1$ if the customer purchases a product, and zero otherwise. Since customers with different profiles respond to products differently, the probability distribution of $Y$ depends on both the product $u \in \mathcal{U}$ that is offered and the customer's profile, and it is given by

$$\mathcal{P}\{Y = 1 | X = x, U = u\} = p_u(x), \quad \forall x \in \mathcal{S}, u \in \mathcal{U},$$

for some function $p_u(x)$. Also, let $q_u(x) = 1 - p_u(x)$. If the customer purchases a product $u \in \mathcal{U}$, a nonnegative reward $R_u \leq R_{\max}$ is obtained.

We assume that the process terminates once the customer purchases a product. Thus, a policy $\pi = (\pi_1, \pi_2, \ldots)$ is a sequence of products to be presented to the customer, where $\pi_t$ is the product that will be shown at time $t$ in the event that the customer did not purchase the products $\pi_1, \ldots, \pi_{t-1}$ during the previous $t - 1$ marketing campaigns.

The objective is to find a sequence of products $\pi = (\pi_1, \pi_2, \ldots)$ that maximizes the expected discounted reward:

$$E_\pi \left[ \sum_{t=1}^{\tau_\pi} \beta^{t-1} R(Y_t, \pi_t) \middle| \phi_0 \right],$$

where $\tau_\pi$ is the first time that the customer purchases a product; $Y_t$ is the customer's response to the product $\pi_t$ at time $t$; $R(Y_t, \pi_t)$ is the reward, given by $R_{\pi_t}$ if $Y_t = 1$ and zero otherwise, and $0 < \beta < 1$ is the discount factor. The discount factor $\beta$ can also be interpreted as the probability that the customer remains engaged in the marketing campaign, given that she was not interested in the product that was just presented to her.

## 3 Intractability of Dynamic Programming

Our POMDP can be converted to a standard Markov decision problem whose states correspond to posterior distributions of $X$. Let $\mathcal{D}_\mathcal{S}$ denote the set of probability distributions on $\mathcal{S}$. For any bounded measurable function $V : \mathcal{D}_\mathcal{S} \to \Re$, let $\bar{T}V : \mathcal{D}_\mathcal{S} \to \Re$ be defined by

$$(\bar{T}V)(\phi) = \max_{u \in \mathcal{U}} \{R_u \mathcal{P}_u\{Y = 1 | \phi\} + \beta V(F_u \phi) \mathcal{P}_u\{Y = 0 | \phi\}\}, \quad \forall \phi \in \mathcal{D}_\mathcal{S},$$

where $\mathcal{P}_u\{Y = 1 | \phi\}$ denotes the probability that the customer will purchase product $u$, provided that the posterior distribution of $X$ is $\phi$. The variable $F_u \phi$ denotes the updated posterior distribution given that the product $u$ was not purchased by the customer, and it is defined by

$$(F_u \phi)(x) = \frac{q_u(x)\phi(x)}{\sum_{z \in \mathcal{S}} q_u(z)\phi(z)}, \quad \forall x \in \mathcal{S}.$$

The optimal value function $V^*$ satisfies the equation: $V^* = \bar{T}V^*$.

The standard dynamic programming algorithm generates a sequence of functions $V_0, V_1, \ldots$ where

$$V_t = \bar{T}V_{t-1}, \quad \forall t \geq 1,$$

with $V_0 = 0$. Because future rewards are discounted, as $t$ increases, $V_t$ becomes a good approximation to $V^*$. Since the space of posterior distributions is continuous, one might consider a grid-based approximation to $V_t$. It is possible to derive a relationship between the number of grid points and the quality of the resulting approximation, but we will not pursue that here. Rather, we just mention that the number of grid points required for an approximation error to be less than $\epsilon$ generally grows exponentially in $|\mathcal{S}|$. Hence, the grid-based approach quickly becomes intractable as $|\mathcal{S}|$ increases.

Although we have encoded state in terms of posterior distribution, one may alternatively encode state in terms of the number of times each product has been rejected by the customer. In this case, the state space at time $t + 1$ is given by $\{(n_1, \ldots, n_{|\mathcal{U}|}) \mid n_i \geq 0 \,\forall i, \sum_i n_i = t\}$, where $n_i$ denotes the number of times that product $i$ was not purchased by the customer during the previous $t$ marketing campaigns. It is not hard to show that the cardinality of this set is given by $\binom{t + |\mathcal{U}| - 1}{|\mathcal{U}| - 1}$, which becomes enormous as $|\mathcal{U}|$ and $t$ grow.

## 4 A Tractable Class of Distributions

Although the general problem is intractable, we will show that for a certain class of distributions, this problem can be solved efficiently. This class of distributions satisfies the following assumption:

**Assumption 1**
(a) For all $u \in \mathcal{U}$,

$$q_u(x) = \prod_{l=1}^{K} f_l^{\zeta_{u,l}}(x), \quad \forall x \in \mathcal{S},$$

where $\zeta_u = (\zeta_{u,1}, \ldots, \zeta_{u,K})$ is the parameter associated with the decision $u$, and $\zeta_{u,l} \geq 0$ for all $u \in \mathcal{U}$ and $l = 1, \ldots, K$. The functions $f_1, \ldots, f_K$ are mappings from $\mathcal{S}$ to $(0, 1]$.
(b) *The vectors $\{\ln f_i(\cdot) \in \Re^{|\mathcal{S}|} | i = 1, \ldots, K\}$ are linearly independent.*

Before we proceed to the analysis, let us motivate our assumptions. Assumption 1(a) simplifies the form of posterior distributions. If $\mathcal{P}\{X = x | a_1, \ldots, a_t\}$ denotes the conditional probability of $X$ given that the customer is not interested in products $a_1, \ldots, a_t$, then

$$\mathcal{P}\{X = x | a_1, \ldots, a_t\}$$
$$= \frac{\phi_0(x) \prod_{r=1}^t q_{a_r}(x)}{\sum_{z \in \mathcal{S}} \phi_0(z) \prod_{r=1}^t q_{a_r}(z)}$$
$$= \frac{\phi_0(x) \prod_{l=1}^K f_l(x)^{\sum_{r=1}^t \zeta_{a_r,l}}}{\sum_{z \in \mathcal{S}} \phi_0(z) \prod_{l=1}^K f_l(z)^{\sum_{r=1}^t \zeta_{a_r,l}}}.$$

The posterior distribution is now characterized by a $K$-dimensional vector $(\sum_r \zeta_{a_r,1}, \ldots \sum_r \zeta_{a_r,K})$. Thus, the class of posterior distributions $\mathcal{D}$ associated with our problem is given by

$$\mathcal{D} = \left\{ g(\cdot, \gamma) \big| \gamma \in \Re_+^K, \right.$$
$$\left. g(x, \gamma) = \frac{\phi_0(x) \prod_{l=1}^K f_l^{\gamma_l}(x)}{\sum_{z \in \mathcal{S}} \phi_0(z) \prod_{l=1}^K f_l^{\gamma_l}(z)}, \ \forall x \in \mathcal{S} \right\}.$$

Recall that in the general problem, posterior distributions of $X$ lie in a $|\mathcal{S}|$-dimensional simplex. With Assumption 1(a), the set of posterior distributions can now be identified with $\Re_+^K$. Thus, the dimension of our state space is reduced to $K$, which can be much smaller than $|\mathcal{S}|$. We assume that for all $i$, $f_i(x) > 0$ for all $x \in \mathcal{S}$. This assumption excludes classes of response functions for which there exists a profile $x \in \mathcal{S}$ such that $q_u(x) = 0$ for all $u$. This unrealistic situation corresponds to a scenario in which customers with a particular profile $x$ will always purchase any product that is offered.

In addition to simplifying the form of posterior distributions, the set of vectors $\{\ln f_i\}$ can be interpreted as a basis used to encode the logarithms of the response functions. In particular, Assumption 1(a) implies that for all $u \in \mathcal{U}$,

$$\ln q_u(x) = \sum_{i=1}^K \zeta_{u,l} \ln f_l(x), \quad \forall x \in \mathcal{S}.$$

Using the fact $\zeta_{u,l} \geq 0$, the above equation implies that the logarithms of the response functions lie in a positive $K$-dimensional cone generated by $\{\ln f_i\}$. Since $\{\ln f_i(\cdot) \in \Re^{|\mathcal{S}|} | i = 1, \ldots, K\}$ is a linearly independent set of vectors by Assumption 1(b), by choosing $K = |\mathcal{S}|$, any response function can be represented in the form required by Assumption 1(a). However, in many cases, these response functions should lie in some lower-dimensional subspace, and thus, $K$ can be much smaller than $|\mathcal{S}|$.

## 5 A Solution Method and Its Analysis

In this section, we will develop an effective approximation method for solving the class of POMDPs that satisfies Assumption 1. As we have noted in the previous section, the class of posterior distributions $\mathcal{D}$ associated with our problem is

$$\mathcal{D} = \left\{ g(\cdot, \gamma) \big| \gamma \in \Re_+^K, \right.$$
$$\left. g(x, \gamma) = \frac{\phi_0(x) \prod_{l=1}^K f_l^{\gamma_l}(x)}{\sum_{z \in \mathcal{S}} \phi_0(z) \prod_{l=1}^K f_l^{\gamma_l}(z)}, \ \forall x \in \mathcal{S} \right\}.$$

Since $\mathcal{D}$ can be identified with $\Re_+^K$, we can also define a value function on $\Re_+^K$. The dynamic programming equation for the optimal value function $J^* : \Re_+^K \to \Re_+$ can now be written as

$$J^*(\gamma) = \max_{u \in \mathcal{U}} \{R_u(1 - H_u(\gamma)) + \beta J^*(\gamma + \zeta_u)H_u(\gamma)\}, \quad (1)$$

for all $\gamma \in \Re_+^K$, where $H_u(\gamma)$ denotes the probability that the customer will not purchase product $u$ given that the posterior distribution of $X$ is $g(\cdot, \gamma)$, and it is given by

$$H_u(\gamma) = \sum_{x \in \mathcal{S}} q_u(x) \frac{\phi_0(x) \prod_{l=1}^K f_l^{\gamma_l}(x)}{\sum_{z \in \mathcal{S}} \phi_0(z) \prod_{l=1}^K f_l^{\gamma_l}(z)}.$$

The derivation of Equation 1 also makes use of the fact that if $g(\cdot, \gamma) \in \mathcal{D}$ is the posterior distribution of $X$, then $g(\cdot, \gamma + \zeta_u)$ is the updated posterior distribution given that the customer rejected product $u$. To facilitate our discussion, let us introduce some notation. Let $\mathcal{B}$ denote the set of bounded measurable functions on $\Re_+^K$, and define $T : \mathcal{B} \to \mathcal{B}$ as follows

$$(TJ)(\gamma) = \max_{u \in \mathcal{U}} \{R_u(1 - H_u(\gamma)) + \beta J(\gamma + \zeta_u) H_u(\gamma)\},$$

for all $J \in \mathcal{B}$ and $\gamma \in \Re_+^K$. Hence, the optimal value function $J^*$ is the fixed point of $T$, i.e. $J^* = TJ^*$. Also, let a sequence of functions $J_0, J_1, \ldots$ be defined by

$$J_t = TJ_{t-1}, \quad \forall t \geq 1,$$

with $J_0 = 0$. Thus, $J_t$ denotes the optimal value function associated with a $t$-time horizon problem.

Before we proceed to the analysis, let us outline the main ideas of our argument. In Section 5.1, we will consider the dynamic programming algorithm for computing approximations to $J^*$, and show that an error bound of the form $\|J^* - J_t\|_\infty \leq \epsilon$ can be obtained with $t \approx O(\ln(1/\epsilon))$. This result enables us to focus our effort on finding good approximations to $J_t$. Since the domain of $J_t$ is $\Re_+^K$, which is unbounded, it is unreasonable to expect a uniformly accurate approximation. Instead, we will only require that our approximation is good over an appropriately bounded region.

Then, in Section 5.2, we will prove that the function $H_u$ is Lipschitz continuous, and as a corollary, that the value function $J_t$ is also Lipschitz continuous. This result motivates grid–based approximations to the value function. In Section 5.3, we define a grid and the corresponding approximation, and show that the performance of the resulting policy is near optimal. Our main result establishes that an $\epsilon$-optimal policy can be generated using $O\left(\left(\frac{1}{\epsilon}\ln\frac{1}{\epsilon}\right)^K\right)$ grid points.

### 5.1 Dynamic Programming

In this section, we study the dynamic programming algorithm for computing approximations to the value function $J^*$. This method is motivated by the following result whose proof follows immediately from the contraction mapping property of $T$, and the fact that $J^*(\gamma) \leq R_{\max}$ for all $\gamma \in \Re_+^K$.

**Lemma 1** *For all $t$,*
$$\|J^* - J_t\|_\infty \leq \beta^t R_{\max}.$$

Let $n(\epsilon)$ be defined by
$$n(\epsilon) = \left\lceil \frac{\ln R_{\max} + \ln \frac{1}{\epsilon}}{\ln \frac{1}{\beta}} \right\rceil.$$

It follows from the above lemma that $\|J^* - J_{n(\epsilon)}\|_\infty \leq \epsilon$. Thus, it suffices to find good approximations to $J_{n(\epsilon)}$. If $\tilde{J}_{n(\epsilon)}$ denotes an approximation to $J_{n(\epsilon)}$, ideally, we would like the error $\left\|J_{n(\epsilon)} - \tilde{J}_{n(\epsilon)}\right\|_\infty$ to be small. However, the domain of $J_{n(\epsilon)}$ is an unbounded set $\Re_+^K$, so finding an approximation that is uniformly accurate may not be possible.

Thus, we will consider an alternative metric that exploits a special feature of our problem formulation. To facilitate our discussion, let $\zeta^*$ be defined by
$$\zeta^* = \max_{u \in \mathcal{U}, l=1,\ldots,K} \zeta_{u,l},$$

and for any positive integer $n$, let
$$\Gamma_n = \left\{\gamma \in \Re_+^K \big| \|\gamma\|_\infty \leq \zeta^* n\right\}.$$

Recall that in our formulation, we start with a prior distribution $\phi_0$ of $X$, which corresponds to $\gamma = 0 \in \Re_+^K$. In addition, it follows from Equation 1 that, at each time period, the value of $\gamma$ can be incremented by at most $\zeta^*$. Since $J_{n(\epsilon)}$ corresponds to the optimal value function for a $n(\epsilon)$-time horizon problem, the "effective domain" of $J_{n(\epsilon)}$ starting at $\gamma = 0$ – the set of possible values of $\gamma$ at time $n(\epsilon)$ – is given by $\Gamma_{n(\epsilon)}$. Therefore, it is natural to restrict the requirement on our approximation to reflect accuracy only over this domain. This motivates the following metric: for any $G \subseteq \Re_+^K$, $J \in \mathcal{B}$, let
$$\|J\|_\infty^G = \sup_{\gamma \in G} |J(\gamma)|.$$

Note that $\|J\|_\infty^G \leq \|J\|_\infty$ for all $J \in \mathcal{B}$. Our goal is to find an approximation $\tilde{J}_{n(\epsilon)}$ to $J_{n(\epsilon)}$ such that the error $\left\|J_{n(\epsilon)} - \tilde{J}_{n(\epsilon)}\right\|_\infty^{\Gamma_{n(\epsilon)}}$ is small.

### 5.2 Lipschitz Condition

In this section, we will show that the function $H_u$ is Lipschitz continuous. This result will enable us to bound error resulting from our approximations. Before we proceed to the statement of this result, let us introduce some notation. Let $M \geq 0$ be defined by

$$\begin{aligned}
M &= K \max_{u \in \mathcal{U}} \left(\max_{x \in \mathcal{S}} q_u(x) - \min_{x \in \mathcal{S}} q_u(x)\right) \\
&\quad \times \max_{i=1,\ldots,K} \left(\max_{x \in \mathcal{S}} \ln f_i(x) - \min_{x \in \mathcal{S}} \ln f_i(x)\right).
\end{aligned}$$

We then have the following result.

**Lemma 2** *For all $\gamma, \gamma' \in \Re_+^K$,*
$$|H_u(\gamma) - H_u(\gamma')| \leq M\|\gamma - \gamma'\|_\infty,$$
*for all $u \in \mathcal{U}$.*

The proof of Lemma 2 makes use of the following result which bounds the derivative of $H_u$. Since the proof of this result consists of simple algebraic manipulations, we refer the reader to our full-length paper for more details.

**Lemma 3** *For all $u \in \mathcal{U}$,*
$$\begin{aligned}
\frac{\partial}{\partial \gamma_i} &H_u(\gamma_1, \ldots, \gamma_K) \\
&= E_\gamma\left[q_u(X) \ln f_i(X)\right] - E_\gamma\left[q_u(X)\right] E_\gamma\left[\ln f_i(X)\right] \\
&= \mathrm{Cov}_\gamma\left(q_u(X), \ln f_i(X)\right)
\end{aligned}$$

*where $E_\gamma[\cdot]$ denotes the expectation with respect to the density $g(\cdot, \gamma)$ defined by*
$$g(x, \gamma) = \frac{\phi_0(x) \prod_{l=1}^K f_l^{\gamma_l}(x)}{\sum_{z \in \mathcal{S}} \phi_0(z) \prod_{l=1}^K f_l^{\gamma_l}(z)}, \quad \forall x \in \mathcal{S}.$$

*Moreover, for all* $u \in \mathcal{U}$,

$$\|(\nabla H_u)(\gamma)\|_1 \leq M, \quad \forall \gamma \in \Re_+^K.$$

Here is the proof of Lemma 2.

*Proof:* Using a standard result that for all $\gamma, \gamma' \in \Re_+^K$,

$$|H_u(\gamma) - H_u(\gamma')|$$
$$\leq \sup_{\delta \in (0,1)} \|(\nabla H_u)(\gamma + \delta(\gamma' - \gamma))\|_1 \|\gamma - \gamma'\|_\infty,$$

it follows from Lemma 3 that $|H_u(\gamma) - H_u(\gamma')| \leq M\|\gamma - \gamma'\|_\infty$. ∎

From the Lipschitz condition of $H_u$, we can also prove that the value function $J_t$ is Lipschitz continuous. This result is stated in the following corollary. Due to the space constraint, the reader is referred to our full-length paper for the proof.

**Corollary 1** *For all* $\gamma, \gamma' \in \Re_+^K$,

$$|J_t(\gamma) - J_t(\gamma')| \leq \frac{(1+\beta)R_{\max}M}{1-\beta} \|\gamma - \gamma'\|_\infty,$$

*for all* $t \geq 0$.

### 5.3 Approximate Value Function

Corollary 1 motivates grid-based dynamic programming techniques. Let $h \in (0,1]$ be a scalar that parameterizes the coarseness of our discretization; we call $h$ the "grid spacing". We start by partitioning the nonnegative half-line $\Re_+$ into a collection $\mathcal{I}_h$ of disjoint subsets. In particular, $\mathcal{I}_h$ consists of sets of the form $[ih, (i+1)h)$ for $i = 0, 1, 2, \ldots$. We then partition $[0,1)^K$ into a collection $\mathcal{I}_h^K$ of subsets defined by

$$\mathcal{I}_h^K = \{I_1 \times \cdots \times I_K | I_i \in \mathcal{I}_h\}.$$

For any $\gamma \in \Re_+^K$, if

$$\gamma \in [i_1h, (i_1+1)h) \times \cdots \times [i_Kh, (i_K+1)h),$$

for some $i_1, \ldots, i_K$, then let $\hat{\gamma}_h = (i_1h, \ldots, i_Kh)$. We will use $\hat{\gamma}_h$ as an approximation to $\gamma$.

We are now ready to define our approximation. Let $\widetilde{T}^h : \mathcal{B} \to \mathcal{B}$ be defined by

$$(\widetilde{T}^h J)(\gamma) = \max_{u \in \mathcal{U}} \{R_u(1 - H_u(\hat{\gamma}_h)) + \beta J(\hat{\gamma}_h + \zeta_u)H_u(\hat{\gamma}_h)\}$$

for all $J \in \mathcal{B}$ and $\gamma \in \Re_+^K$. We should note that in order to compute the function $\widetilde{T}^h J$, it suffices to compute $\widetilde{T}^h J$ only at the grid points. Since we are only interested in approximating $J_{n(\epsilon)}$ on the set $\Gamma_{n(\epsilon)}$, the maximal value of $\gamma$ that we need to consider is

$$\zeta^* n(\epsilon) + \sum_{i=0}^{n(\epsilon)-1} \zeta^* = 2\zeta^* n(\epsilon).$$

Thus, we only need to define our approximate value functions on subsets of $\Gamma_{2n(\epsilon)}$. So, let a sequence of approximate value functions $\tilde{J}_0^h, \ldots, \tilde{J}_{n(\epsilon)}^h$, where $\tilde{J}_t^h : \Gamma_{2n(\epsilon)-t} \to \Re_+$, be defined by

$$\tilde{J}_t^h(\gamma) = \left(\widetilde{T}^h \tilde{J}_{t-1}^h\right)(\gamma), \quad \forall \gamma \in \Gamma_{2n(\epsilon)-t},\ t \leq n(\epsilon),$$

with $\tilde{J}_0^h = 0$. The following theorem shows that our approximate value function is close to the true value function. The proof of this theorem will be given in Section 5.4.

**Theorem 1** *For all* $t \leq n(\epsilon)$,

$$\left\|J_t - \tilde{J}_t^h\right\|_\infty^{\Gamma_{2n(\epsilon)-t}} \leq \frac{(1+\beta)R_{\max}M}{(1-\beta)^2} h.$$

The above result suggests that the performance of a greedy policy generated from the approximate value function should also be close to optimal. Our approximate policy $\hat{\mu}^h = \left\{\hat{\mu}_1^h, \ldots, \hat{\mu}_{n(\epsilon)}^h\right\}$, where $\hat{\mu}_t^h : \Gamma_{2n(\epsilon)-t} \to \mathcal{U}$, is defined by

$$T_{\hat{\mu}_t^h} \tilde{J}_{t-1}^h = T\tilde{J}_{t-1}^h, \quad \forall t \geq 1,$$

where for any decision rule $\mu$, $T_\mu$ is defined by

$$(T_\mu J)(\gamma) = R_{\mu(\gamma)}\left(1 - H_{\mu(\gamma)}(\gamma)\right) + \beta J\left(\gamma + \zeta_{\mu(\gamma)}\right) H_{\mu(\gamma)}(\gamma),$$

for all $J \in \mathcal{B}$ and $\gamma \in \Re_+^K$. If $\hat{J}_t^h$ denotes the expected reward for a $t$-time horizon problem under the policy $\hat{\mu}^h$, then $\hat{J}_t^h$ satisfies the following equation:

$$\hat{J}_t^h = T_{\hat{\mu}_t^h} \hat{J}_{t-1}^h, \quad \forall t \geq 1,$$

with $\hat{J}_0^h = 0$. The following theorem asserts that the performance of our greedy policy is close to the optimal performance.

**Theorem 2** *For all* $t \leq n(\epsilon)$,

$$\left\|J_t - \hat{J}_t^h\right\|_\infty^{\Gamma_{2n(\epsilon)-t}} \leq \frac{2\beta(1+\beta)R_{\max}M}{(1-\beta)^3} h$$

The proof of Theorem 2 makes use of the following lemma which shows that the operators $T$, $\widetilde{T}^h$, and $T_\mu$ are contraction mappings. This lemma follows immediately from the definition of these operators, and we omit the proof.

**Lemma 4** *For any* $G_1, G_2 \in \mathcal{B}$, *and* $\mu : \Re_+^K \to \mathcal{U}$,

$$\|TG_1 - TG_2\|_\infty^{\Gamma_n} \leq \beta \|G_1 - G_2\|_\infty^{\Gamma_{n+1}},$$
$$\left\|\widetilde{T}^h G_1 - \widetilde{T}^h G_2\right\|_\infty^{\Gamma_n} \leq \beta \|G_1 - G_2\|_\infty^{\Gamma_{n+1}},$$
$$\|T_\mu G_1 - T_\mu G_2\|_\infty^{\Gamma_n} \leq \beta \|G_1 - G_2\|_\infty^{\Gamma_{n+1}}$$

*for all* $n \geq 0$.

Here is the proof of Theorem 2.
*Proof:* For any $t \leq n(\epsilon)$,

$$\begin{aligned}\left\|J_t - \hat{J}_t^h\right\|_\infty^{\Gamma_{2n(\epsilon)-t}} &= \left\|TJ_{t-1} - T_{\hat{\mu}_t^h}\hat{J}_{t-1}^h\right\|_\infty^{\Gamma_{2n(\epsilon)-t}} \\ &\leq \left\|TJ_{t-1} - T_{\hat{\mu}_t^h}\tilde{J}_{t-1}^h\right\|_\infty^{\Gamma_{2n(\epsilon)-t}} \\ &+ \left\|T_{\hat{\mu}_t^h}\tilde{J}_{t-1}^h - T_{\hat{\mu}_t^h}J_{t-1}\right\|_\infty^{\Gamma_{2n(\epsilon)-t}} \\ &+ \left\|T_{\hat{\mu}_t^h}J_{t-1} - T_{\hat{\mu}_t^h}\hat{J}_{t-1}^h\right\|_\infty^{\Gamma_{2n(\epsilon)-t}}\end{aligned}$$

By definition of $\hat{\mu}_t^h$, we have

$$T_{\hat{\mu}_t^h}\tilde{J}_{t-1}^h = T\tilde{J}_{t-1}^h,$$

which implies that

$$\left\|TJ_{t-1} - T_{\hat{\mu}_t^h}\tilde{J}_{t-1}^h\right\|_\infty^{\Gamma_{2n(\epsilon)-t}} \leq \beta \left\|J_{t-1} - \tilde{J}_{t-1}^h\right\|_\infty^{\Gamma_{2n(\epsilon)-t+1}},$$

where the inequality follows from Lemma 4. Similarly, Lemma 4 implies that

$$\left\|T_{\hat{\mu}_t^h}\tilde{J}_{t-1}^h - T_{\hat{\mu}_t^h}J_{t-1}\right\|_\infty^{\Gamma_{2n(\epsilon)-t}} \leq \beta \left\|\tilde{J}_{t-1}^h - J_{t-1}\right\|_\infty^{\Gamma_{2n(\epsilon)-t+1}}$$

and

$$\left\|T_{\hat{\mu}_t^h}J_{t-1} - T_{\hat{\mu}_t^h}\hat{J}_{t-1}^h\right\|_\infty^{\Gamma_{2n(\epsilon)-t}} \leq \beta \left\|J_{t-1} - \hat{J}_{t-1}^h\right\|_\infty^{\Gamma_{2n(\epsilon)-t+1}}$$

Hence, it follows from Theorem 1 that

$$\begin{aligned}&\left\|J_t - \hat{J}_t^h\right\|_\infty^{\Gamma_{2n(\epsilon)-t}} \\ &\leq \frac{2\beta(1+\beta)R_{\max}M}{(1-\beta)^2}h + \beta\left\|J_{t-1} - \hat{J}_{t-1}^h\right\|_\infty^{\Gamma_{2n(\epsilon)-t+1}}\end{aligned}$$

Since $J_0 = \hat{J}_0^h = 0$, the above recursion implies that

$$\left\|J_t - \hat{J}_t^h\right\|_\infty^{\Gamma_{2n(\epsilon)-t}} \leq \frac{2\beta(1+\beta)R_{\max}M}{(1-\beta)^3}h$$

for all $t$. ∎

Since

$$\begin{aligned}&\left\|J^* - \hat{J}_{n(\epsilon)}^h\right\|_\infty^{\Gamma_{n(\epsilon)}} \\ &\leq \left\|J^* - J_{n(\epsilon)}\right\|_\infty^{\Gamma_{n(\epsilon)}} + \left\|J_{n(\epsilon)} - \hat{J}_{n(\epsilon)}^h\right\|_\infty^{\Gamma_{n(\epsilon)}} \\ &\leq \left\|J^* - J_{n(\epsilon)}\right\|_\infty + \left\|J_{n(\epsilon)} - \hat{J}_{n(\epsilon)}^h\right\|_\infty^{\Gamma_{n(\epsilon)}},\end{aligned}$$

the following corollary follows immediately from Lemma 1 and Theorem 2.

**Corollary 2**

$$\left\|J^* - \hat{J}_{n(\epsilon)}^h\right\|_\infty^{\Gamma_{n(\epsilon)}} \leq \epsilon + \frac{2\beta(1+\beta)R_{\max}M}{(1-\beta)^3}h$$

The above corollary shows that in order for our approximation to be within $2\epsilon$ of the optimal value function, we need to approximate $J_{n(\epsilon)}$ using a grid spacing of

$$h = \frac{\epsilon(1-\beta)^3}{2\beta(1+\beta)R_{\max}M}.$$

Since our approximate value functions are defined on subsets of $\Gamma_{2n(\epsilon)}$, the maximum number of grid points is of order

$$O\left(\left(\zeta^* M \frac{2n(\epsilon)}{\epsilon}\right)^K\right) = O\left(\left(\zeta^* M \frac{1}{\epsilon} \ln \frac{1}{\epsilon}\right)^K\right),$$

where the equality follows from the definition of $n(\epsilon)$, and we have ignored the constants $R_{\max}$ and $\beta$ since these two variables generally do not scale with the problem size. Let us first note the dependence of the number of grid points on the error tolerance $\epsilon$. Note that

$$\frac{1}{\epsilon}\ln\frac{1}{\epsilon} = O\left(\frac{1}{\epsilon^p}\right),$$

for all $p > 1$. Since any other grid-based approximation method would require at least $O((1/\epsilon)^K)$ grid points, the number of grid points used by our approximation method is comparable to even the best grid-based approximation technique.

Let us now consider the dependence of the number of grid points on the constant $M$. Recall that

$$M = C_1 \times C_2,$$

where

$$C_1 = K \max_{u \in \mathcal{U}} \left(\max_{x \in \mathcal{S}} q_u(x) - \min_{x \in \mathcal{S}} q_u(x)\right),$$

and

$$C_2 = \max_{i=1,\ldots,K}\left(\max_{x \in \mathcal{S}} \ln f_i(x) - \min_{x \in \mathcal{S}} \ln f_i(x)\right).$$

The constant $C_1$ represents the maximum variability in the response functions, relative to the number of basis functions $K$. At first glance, it seems that $C_1$ should increase proportionally with $K$. This would imply that the number of grid points would quickly become intractable as the problem size increases and we need more basis functions. However, we believe that, in most cases, $C_1$ will remain bounded even when the problem size increases. As an example, consider a situation where

$$f_i(x) \leq \alpha < 1, \quad \forall x \in \mathcal{S}, i = 1, \ldots, K,$$

and

$$\underline{\zeta} = \min_{u \in \mathcal{U}, i=1,\ldots,K} \zeta_{u,l} > 0.$$

In that case,
$$q_u(x) = \prod_{l=1}^{K} f_l^{\zeta_{u,l}}(x) \leq \left(\alpha^{\underline{\zeta}}\right)^K, \quad \forall x \in \mathcal{S}, u \in \mathcal{U},$$
where $0 < \alpha^{\underline{\zeta}} < 1$. For any function $f_\theta : \Re_+ \to \Re_+$, $0 < \theta < 1$, defined by
$$f_\theta(w) = w\theta^w, \quad \forall w \in \Re_+,$$
one can verify that
$$\max_{w \in \Re_+} f_\theta(w) = f_\theta(w)|_{w=\log_\theta(\frac{1}{e})} = \frac{1}{e} \log_\theta\left(\frac{1}{e}\right),$$
where $e$ denotes the base of the natural logarithm, and $\log_\theta(\cdot)$ denotes the logarithm with base $\theta$. Therefore,
$$K \max_{u \in \mathcal{U}} \left(\max_{x \in \mathcal{S}} q_u(x) - \min_{x \in \mathcal{S}} q_u(x)\right) \leq \frac{1}{e} \log_{\alpha^{\underline{\zeta}}}\left(\frac{1}{e}\right),$$
which shows that $C_1$ is bounded above by a constant that does not increase with $K$.

The constant $C_2$ denotes the variability in our basis functions $\{\ln f_i\}$. From Assumption 1(a), we have
$$\ln q_u(x) = \sum_{l=1}^{K} \zeta_{u,l} \ln f_l(x), \quad \forall x \in \mathcal{S}, u \in \mathcal{U}.$$
Thus, the logarithms of our response functions lie in a positive cone generated by $\{\ln f_i\}$. If $C_2$ is large, then our basis functions can represent a larger class of response functions. Hence, the number of required grid points increases proportionally with the representation capability of our basis functions. In the extreme case where $C_2 = 0$, $f_i$ is a constant function for all $i$. In that case, the only response functions that can be represented using this basis are the ones where $q_u(x)$ is constant for all $x \in \mathcal{S}$. In most cases, we expect that
$$f_i(x) \geq \delta, \quad \forall x \in \mathcal{S}, i = 1, \ldots, K,$$
for some $\delta > 0$. Under this condition, we have
$$\max_i \left(\max_{x \in \mathcal{S}} \ln f_i(x) - \min_{x \in \mathcal{S}} \ln f_i(x)\right) \leq \ln(1/\delta),$$
which shows that $C_2$ does not increase with the problem size.

Finally, let us consider the dependence of the number of grid points on the constant $\zeta^*$, which was defined in Section 5.1 as
$$\zeta^* = \max_{u \in \mathcal{U}, l=1,\ldots,K} \zeta_{u,l}.$$
We believe that, in most cases, $\zeta^*$ will remain bounded even when the problem size increases and more products are considered. As an example, consider a situation where
$$0 < \nu \leq q_u(x), \quad \forall x \in \mathcal{S}, u \in \mathcal{U}.$$

Under this condition, if $\zeta^* = \zeta_{u^*, l^*}$ for some $u^* \in \mathcal{U}$ and $l^* \in \{1, \ldots, K\}$, then
$$\nu \leq q_{u^*}(x) = f_{l^*}^{\zeta^*}(x) \prod_{l \neq l^*} f_l^{\zeta_{u^*,l}}(x), \quad \forall x \in \mathcal{S},$$
which implies that
$$\zeta^* \leq \frac{\ln \nu - \sum_{l \neq l^*} \zeta_{u^*, l} \ln f_l(x)}{\ln f_{l^*}(x)} \leq \frac{\ln(1/\nu)}{\ln(1/f_{l^*}(x))}.$$
From our discussion on the scalability of the constant $C_1$, we expect that $f_l(x) \leq \alpha < 1$, $\forall x \in \mathcal{S}, l = 1, \ldots, K.$, which implies that
$$\zeta^* \leq \frac{\ln(1/\nu)}{\ln(1/\alpha)}.$$
The above inequality shows that $\zeta^*$ does not increase with the problem size.

### 5.4 Proof of Theorem 1

*Proof:* For any $t \leq n(\epsilon)$,
$$\left\|J_t - \tilde{J}_t^h\right\|_\infty^{\Gamma_{2n(\epsilon)-t}} = \sup_{\gamma \in \Gamma_{2n(\epsilon)-t}} \left|J_t(\gamma) - \tilde{J}_t^h(\gamma)\right|.$$
It follows from the definition of $J_t$ and $\tilde{J}_t^h$ that
$$\left|J_t(\gamma) - \tilde{J}_t^h(\gamma)\right|$$
$$= \left|(TJ_{t-1})(\gamma) - (\tilde{T}^h \tilde{J}_{t-1}^h)(\gamma)\right|$$
$$\leq \max_{u \in \mathcal{U}} \Big\{ R_u \left|H_u(\gamma) - H_u(\hat{\gamma}_h)\right| + $$
$$\beta \left|J_{t-1}(\gamma + \zeta_u)H_u(\gamma) - \tilde{J}_{t-1}^h(\hat{\gamma}_h + \zeta_u)H_u(\hat{\gamma}_h)\right| \Big\}$$
$$\leq \max_{u \in \mathcal{U}} \Big\{ R_{\max} M \|\gamma - \hat{\gamma}_h\|_\infty + $$
$$\beta \left|J_{t-1}(\gamma + \zeta_u)H_u(\gamma) - \tilde{J}_{t-1}^h(\hat{\gamma}_h + \zeta_u)H_u(\hat{\gamma}_h)\right| \Big\}$$
where the last inequality follows from Lemma 2. However,
$$\left|J_{t-1}(\gamma + \zeta_u)H_u(\gamma) - \tilde{J}_{t-1}^h(\hat{\gamma}_h + \zeta_u)H_u(\hat{\gamma}_h)\right|$$
$$\leq J_{t-1}(\gamma + \zeta_u) \left|H_u(\gamma) - H_u(\hat{\gamma}_h)\right| + $$
$$H_u(\hat{\gamma}_h) \left|J_{t-1}(\gamma + \zeta_u) - J_{t-1}(\hat{\gamma}_h + \zeta_u)\right| + $$
$$H_u(\hat{\gamma}_h) \left|J_{t-1}(\hat{\gamma}_h + \zeta_u) - \tilde{J}_{t-1}^h(\hat{\gamma}_h + \zeta_u)\right|$$
$$\leq R_{\max} M \|\gamma - \hat{\gamma}_h\|_\infty + $$
$$\frac{(1+\beta)R_{\max}M}{1-\beta} \|\gamma - \hat{\gamma}_h\|_\infty + $$
$$\left|J_{t-1}(\hat{\gamma}_h + \zeta_u) - \tilde{J}_{t-1}^h(\hat{\gamma}_h + \zeta_u)\right|$$
$$\leq \frac{2R_{\max}M}{1-\beta} \|\gamma - \hat{\gamma}_h\|_\infty + $$
$$\left\|J_{t-1} - \tilde{J}_{t-1}^h\right\|_\infty^{\Gamma_{2n(\epsilon)-t+1}},$$

where the next to last inequality follows from Lemma 2 and Corollary 1, along the fact that $J_{t-1}(\gamma) \leq R_{\max}$ and $H_u(\gamma) \leq 1$ for all $\gamma \in \Re_+^K$. The last inequality follows from the fact that $\gamma \in \Gamma_{2n(\epsilon)-t}$ and $\hat{\gamma}_h \leq \gamma$. Thus, putting everything together and using the fact that $\|\gamma - \hat{\gamma}_h\|_\infty \leq h$, we obtain

$$\left\| J_t - \tilde{J}_t^h \right\|_\infty^{\Gamma_{2n(\epsilon)-t}} \leq \frac{(1+\beta)R_{\max}M}{1-\beta}h + \beta \left\| J_{t-1} - \tilde{J}_{t-1}^h \right\|_\infty^{\Gamma_{2n(\epsilon)-t+1}}.$$

Since $J_0 = \tilde{J}_0^h = 0$, the above recursion implies that

$$\left\| J_t - \tilde{J}_t^h \right\|_\infty^{\Gamma_{2n(\epsilon)-t}} \leq \frac{(1+\beta)R_{\max}M}{(1-\beta)^2}h,$$

for all $t \leq n(\epsilon)$. ∎

## 6 A Motivating Application

In this section, we consider an application of our model to a problem in targeted advertising. Consider a retailer who would like to develop a marketing campaign to attract new customers. Let $\mathcal{U}$ denote the set of available products. Assume the we have demographic information and purchase history on our existing customers. Let $X_1, \ldots, X_n$ denote demographic variables that are deemed to be good predictors of a customer's propensity to buy a product. Although we do not have demographic information on new customers that we would like to attract through our marketing campaign, we can exploit the information available in our existing database. For instance, we might assume that the demographic characteristics of the population that is targeted by our campaign has the same distribution as that of our previous customers.

Let us assume without loss of generality that the $X_i$'s are binary variables. We can think of $X = (X_1, \ldots, X_n)$ as a random variable that represents a customer's profile. Then, this problem falls within the framework of our model. In this case, the set $S$ of possible values of $X$ has cardinality $2^n$. Thus, a solution via dynamic programming requires us to compute a value function over a $2^n$-dimensional space of belief states, which quickly becomes intractable as $n$ increases.

However, if the probability that a customer will purchase a product given her demographic characteristics exhibits a noisy-OR structure [5], then this problem becomes tractable. The noisy-OR structure assumes that different demographic factors independently act to influence a customer's purchasing decision. This form of conditional independence has been used successfully to model problems, for example, in medical diagnosis [2]. The noisy-OR model leads to a function of the form

$$q_u(x) = \prod_{l=1}^{n}(1-d_l)^{x_l\zeta_{u,l}}, \quad \forall x \in \{0,1\}^n.$$

The parameter $d_l$ can be interpreted as the baseline probability that the demographic characteristic $X_l$ leads the customer to purchase an arbitrary product, and the parameter $\zeta_{u,l}$ represents the deviation from the baseline probability associated with the product $u \in \mathcal{U}$. As $\zeta_{u,l}$ decreases, the probability that the demographic characteristic $X_l$ will lead the customer to purchase product $u$ also decreases.

Note that the response function takes the form required by Assumption 1(a). Thus, this problem can be solved using a grid on an $n$-dimensional space. This offers a significant reduction in the amount of computation since the dimension of the state space is reduced from $2^n$ to $n$.

## 7 Conclusion

We studied a POMDP that models a class of sequencing problems. Although the general problem is intractable, we show that for a certain class of distributions, the problem can be solved efficiently. Our current research focuses on extending the results to broader classes of problems. We also hope to further explore applications of the model developed in this paper.

### Acknowledgments

This research was supported by NSF CAREER Grant ECS-9985229, by the ONR under Grant MURI N00014-00-1-0637, and by a Stanford Graduate Fellowship.